%% file: paper.tex
\documentclass[11pt,twocolumn]{scrartcl}

\usepackage{casa_conf}	
\usepackage{graphicx}	
\usepackage{verbatim}
\usepackage{booktabs}
\usepackage{amssymb}
\usepackage{amsmath}
\usepackage{algorithm}
\usepackage{algpseudocode}
\usepackage[numbers, sort]{natbib}
\usepackage{xcolor}
\usepackage[normalem]{ulem}
\usepackage{hyperref}

\title{Diverse 3D Human Pose Generation in Scenes based on Decoupled Structure}

\author{Bowen Dang\\
        Xi'an Jiaotong University\\
        dangbowen.bell@gmail.com\\
        \and
        Xi Zhao\thanks{Corresponding Author}\\
        Xi'an Jiaotong University\\
        xi.zhao@mail.xjtu.edu.cn\\
        }

\begin{document}

\maketitle

\begin{abstract}
This paper presents a novel method for generating diverse 3D human poses in scenes with semantic control. Existing methods heavily rely on the human-scene interaction dataset, resulting in a limited diversity of the generated human poses. To overcome this challenge, we propose to decouple the pose and interaction generation process. Our approach consists of three stages: pose generation, contact generation, and putting human into the scene. We train a pose generator on the human dataset to learn rich pose prior, and a contact generator on the human-scene interaction dataset to learn human-scene contact prior. Finally, the placing module puts the human body into the scene in a suitable and natural manner. The experimental results on the PROX dataset demonstrate that our method produces more physically plausible interactions and exhibits more diverse human poses. Furthermore, experiments on the MP3D-R dataset further validates the generalization ability of our method.
\end{abstract}
\linebreak
\linebreak

\input{sections/symbol}
\input{sections/1-introduction}
\input{sections/2-related_work}

\input{sections/3-method}
\input{sections/4-experiment}

\input{sections/5-conclusion}
\input{sections/6-acknowledgments}

\bibliographystyle{unsrtnat}
\bibliography{reference}

\end{document}

%% file: sections/symbol.tex

\newcommand{\shape}{ \beta }
\newcommand{\pose}{ \theta }
\newcommand{\face}{ \psi }
\newcommand{\trans}{ t }

\newcommand{\bpose}{ \hat{\theta_b} }
\newcommand{\pbpose}{ \theta_b }

\newcommand{\contact}{ \hat{f} }
\newcommand{\pcontact}{ f }

\newcommand{\pg}{ L_{\mathrm{Pose}}}
\newcommand{\cg}{ L_{\mathrm{Contact}}}

\newcommand{\mterm}{ L_m }
\newcommand{\vterm}{ L_v }
\newcommand{\jterm}{ L_j }
\newcommand{\klterm}{ L_{\mathrm{kl}}}

\newcommand{\recterm}{ L_{\mathrm{rec}}}

\newcommand{\rterm}{ E_r }
\newcommand{\wcterm}{ E_{\mathrm{wc}} }
\newcommand{\vpterm}{ E_{\mathrm{vp}} }

\newcommand{\mweight}{ \lambda_m }
\newcommand{\vweight}{ \lambda_v }
\newcommand{\jweight}{ \lambda_j }
\newcommand{\klweight}{ \lambda_{\mathrm{kl}}}

\newcommand{\recweight}{ \lambda_{\mathrm{rec}}}

\newcommand{\rweight}{ \lambda_r }
\newcommand{\wcweight}{ \lambda_{\mathrm{wc}} }
\newcommand{\vpweight}{ \lambda_{\mathrm{vp}} }

\newcommand{\vertices}{ V_b }
\newcommand{\svertices}{ V_s }
\newcommand{\opoints}{ P_o }
\newcommand{\cvertices}{ V_c }
\newcommand{\internal}{ P_{\mathrm{int}} }

\newcommand{\faces}{ F_b }
\newcommand{\mean}{ \mu }
\newcommand{\var}{ \sigma }

%% file: sections/1-introduction.tex
\section{Introduction}

Generating natural and diverse human poses is a challenging research problem with wide-ranging applications, such as AR/VR, computer games, and generating training data for vision and graphics tasks. Most methods focus on generating human poses without considering scene constraints \cite{ACTOR, TEMOS}. With the development of the human-scene interaction dataset \cite{PROX, Egobody}, many recent methods \cite{PSI, PLACE, POSA, COINS, Narrator, GenZI} have been dedicated to generating human poses in scenes. In this case, the generated human body must be coherent with the scene's semantic and geometric features to form reasonable spatial relationships with the scene. Our work belongs to this category as well. We focus on generating diverse 3D human poses in scenes with semantic control.

Existing methods \cite{PSI, PLACE} can generate human poses given the surrounding 2D or 3D scene information. However, these methods lack controllability and require manual labor to search for desired interaction types. Recently, some methods have incorporated semantic control into pose generation \cite{COINS, Narrator, GenZI}. Despite their capability to generate semantically plausible human poses, these methods heavily rely on the specific human-scene interaction dataset, and it is challenging for these methods to create diverse human poses that never appear in the interaction dataset. In summary, current works have difficulty generating controllable and diverse human poses while maintaining natural interactions.

We propose a new system based on the decoupled structure to deal with the above problems. Our main idea is to decouple the pose and interaction process. This design enables us to learn a rich pose prior on the large human dataset such as AMASS \cite{AMASS}, so as to minimize the reliance on the human-scene interaction dataset. We generate specified interactions based on the given word instruction that provides the action and object type. The action type controls the generated pose, while the object type controls the interaction mode. By separating the pose and interaction generation modules, our system can produce more diverse human body poses and ensure reasonable interactions.

Our method includes three stages. The first stage is to generate the desired human body pose. We train the pose generator on the human dataset, which is easy to get and has rich annotations. So we can generate various poses to enrich the results. The second stage involves the generation of a contact feature map for the previously generated human body. We train the contact generator on the human-scene interaction dataset to capture the human-scene contact prior. The final stage is to place the human body in the scene. In this stage, we first select initial positions to put the human body. Then, we propose an effective physical feasibility test to remove unsuitable initial results. Finally, we optimize the human body pose to make the result look more natural.

In summary, our contributions are as follows:

\begin{enumerate}
    \item A multi-stage generation framework that decouples the pose and interaction generation process;
    \item A simple yet effective physical feasibility test module to ensure the physical feasibility of the generated results;
    \item We demonstrate that our method can generate more physically plausible interactions with more diverse human poses in scenes compared to other methods.
\end{enumerate}

%% file: sections/2-related_work.tex
\section{Related Work}

\textbf{3D Human Pose Generation in Scenes}: Generating 3D human poses in scenes has been a challenging research problem, and various methods based on different settings have been proposed. Some methods generate the human poses conditioned solely on the scene feature \cite{PSI, PLACE, POSA, PoseGuided}. \citet{PSI} propose to extract the scene feature from the scene semantic segmentation and depth map, and use the feature to generate semantically plausible human poses. \citet{PLACE} model the proximal relationship between the human body and the scene using BPS \cite{BPS} feature. \citet{POSA} propose a body-centric representation that encodes geometric and semantic information of the given human body, and use it to guide the search for the most likely position in the scene. \citet{PoseGuided} introduce a geometric alignment term in the optimization stage to ensure more natural contact with the scene. Recently, some methods introduce semantic control to make the generation process controllable \cite{COINS, Narrator}. \citet{COINS} first generate a plausible human pelvis location and then generate the human body. The proposed method can support atomic and compositional interactions. \citet{Narrator} propose to reason the relationship of the scene structure and use it to generate the desired human poses according to textual descriptions. Our method tackles the same task with COINS \cite{COINS} and we focus on improving the diversity of the human poses by decoupling the pose and interaction generation process.

\textbf{Human-Scene Interaction Representation}: Human-scene interaction has also been widely studied in other tasks such as human or/and scene reconstruction \cite{OcclusionHPR, MOVER} and scene generation \cite{Pigraphs, SUMMON}. \citet{OcclusionHPR} propose to estimate the possible region the human body can be positioned from the scene information, and use it to constrain the human pose for human pose reconstruction under severe occlusions. \citet{MOVER} propose a framework to reconstruct the plausible scene layout from the video and human movement. \citet{Pigraphs} propose to learn a joint distribution of human poses and object arrangements and generate a plausible interaction by sampling from the distribution. \citet{SUMMON} propose to generate the scene layout from the human motions. We focus on generating diverse human poses in scenes. Our method incorporates semantic information while generating the body pose and contact feature.

%% file: sections/3-method.tex
\section{Method}

Our goal is to generate 3D human poses in scenes. Our main idea is to decouple the pose and interaction generation process to minimize the reliance on the human-scene interaction dataset. In this section, we present details of our method.


\begin{figure*}[t]
  \centering
  \includegraphics[width=\linewidth]{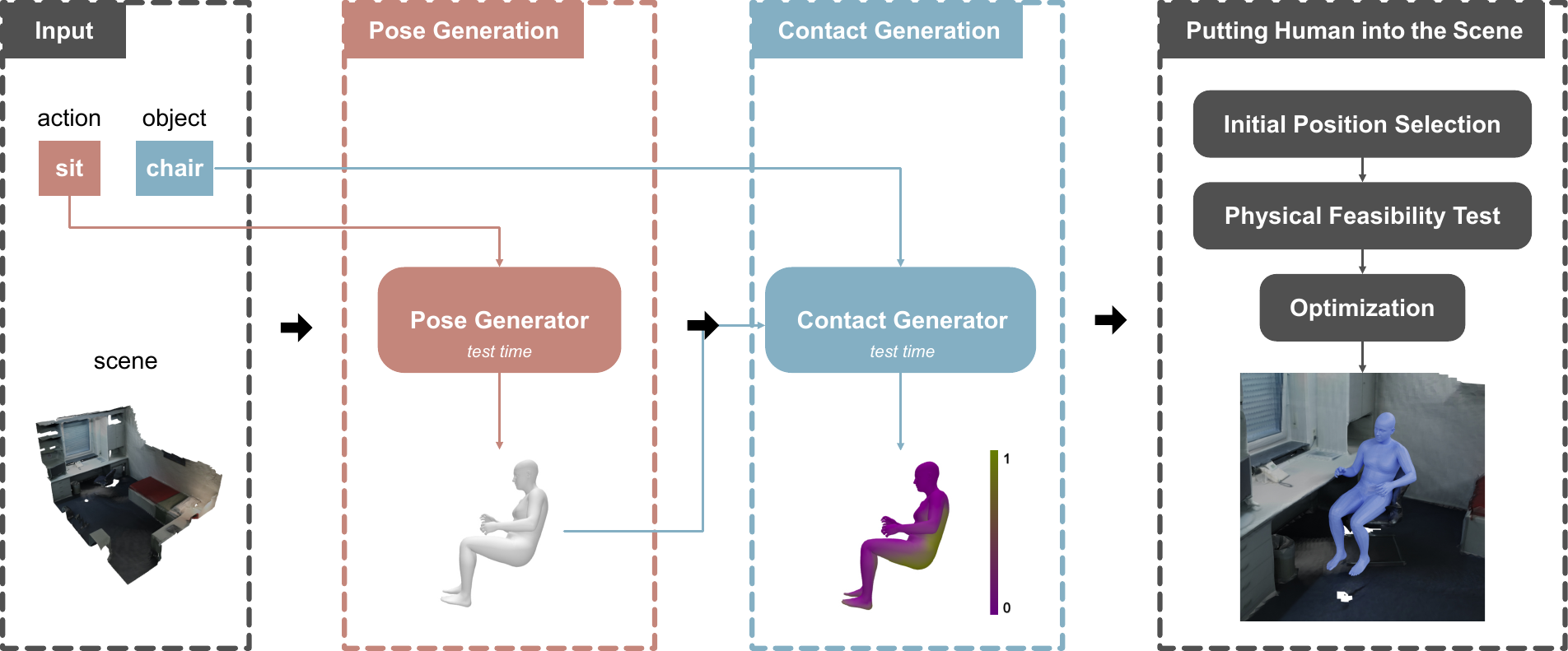}
  \caption{The overview of our method. The input are an action-object pair and the scene mesh. The output is the human body mesh placed in the scene. In the first stage, we generate a desired human body model using the pose generator. In the second stage, we generate the contact feature map for the human body mesh using the contact generator. Finally, we put the human body into the scene. The last stage can be further divided into three sub-stages, including initial position selection, physical feasibility test, and optimization.}
  \label{figure:overview}
\end{figure*}

\subsection{Overview}

Our method takes an action-object pair and the scene mesh as input and generates the human body mesh placed in the scene as output. As shown in Figure \ref{figure:overview}, our method comprises three stages. In the first stage, we use a pose generator to produce a desired human body model. In the second stage, we employ a contact generator to create a contact feature map for the human body mesh. Finally, we place the human body in the scene, which involves three sub-stages: initial position selection, physical feasibility test, and optimization.

We adopt the SMPL-X human body model \cite{SMPLX} to represent the human, which is a differentiable function that takes shape parameters $\shape$, pose parameters $\pose$, facial expression parameters $\face$, and global translation $t$ as input. The output is a human body mesh $M_b = (\vertices, \faces)$, comprising vertices $\vertices$ and faces $\faces$. Throughout our method, we focus on the human body pose $\pbpose$ while keeping the other parameters constant. We assume the scene has semantic and instance segmentation, allowing us to search for plausible positions around the object to place the human body.

\subsection{Pose Generator}

Existing methods typically employ a generator trained on the human-scene interaction dataset to directly generate human poses in scenes \cite{PSI, PLACE, Narrator}. However, the limited availability of interaction dataset \cite{PROX, Egobody} and the restricted range of human poses they contain hinder the diversity of the generated results. To alleviate this dependence on the interaction dataset, we propose an action-conditioned pose generator that is independent of other stages. By training the pose generator on the readily available human dataset with rich annotations, it can learn a rich pose prior, ultimately enhancing the diversity of the generated human poses.

\begin{figure}[t]
  \centering
  \includegraphics[width=0.8\linewidth]{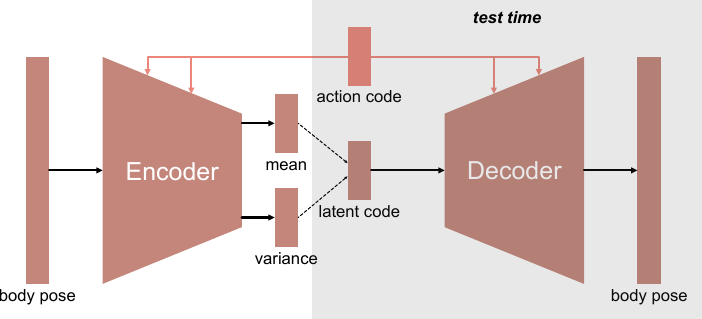}
  \caption{The network structure of the pose generator. The input and output are both the body pose. The action code serves as the conditional input to control the generated body pose.}
  \label{figure:pose_generator}
\end{figure}

As illustrated in Figure \ref{figure:pose_generator}, our approach employs a VAE-based \cite{VAE, SMPLX} network architecture to generate body poses from given action types. The input body pose $\bpose \in R^{63}$ is encoded using a multi-layer perceptron (MLP) encoder, which outputs the mean $\mean \in R^{64}$ and variance $\var \in R^{64}$ of the Gaussian distribution that $\bpose$ belongs to. Subsequently, we obtain the reconstructed body pose $\pbpose \in R^{63}$ by feeding the sampled latent code $z \in R^{64}$ into an MLP decoder. The action code $c_a \in R^{3}$ is defined as the one-hot code representing the action type. It serves as a conditional input to control the generated body pose and is concatenated with the hidden features of both the encoder and decoder. During the testing stage, we only use the decoder part to generate the body pose.

The training loss for the pose generator can be formulated as follows:

\begin{gather}
    \pg = \mweight \mterm + \vweight \vterm + \jweight \jterm + \klweight \klterm \\
    \mterm = \mathrm{Geodesic}(M(\pbpose), M(\bpose)) \\
    \vterm = \|V(\pbpose) - V(\bpose)\|_1 \\
    \jterm = \|J(\pbpose) - J(\bpose)\|_1 \\
    \klterm = \mathrm{KL}(q(Z|\bpose) \| N(0, I))
\end{gather}

\noindent $\mterm$ denotes the pose reconstruction loss and is calculated as the Geodesic distance of rotation matrixs representing the input and reconstructed body pose \cite{Geodesic}. $M(\cdot)$ denotes the function to transform the body pose from axis angle format to rotation matrix format. $\vterm/\jterm$ denotes the vertices/joints reconstruction loss and is calculated as the mean L1 distance between the input and reconstructed body vertices/joints. $V(\cdot)/J(\cdot)$ denotes the function to get the human body vertices/joints. $\klterm$ denotes the KL divergence and is used to encourage the latent space distribution to be close to a prior distribution such as the standard normal distribution. $\lambda_{*}$ denotes the weight for each loss term.

\subsection{Contact Generator}

After obtaining the desired human body model, we need to place it in the scene and ensure that it interacts naturally with the environment. Since most interactions involve contact, the key challenge is to ensure that the human body makes reasonable contact with the scene. To achieve this, we propose learning the human-scene contact prior from the human-scene interaction dataset and using the prior information to guide the human body to contact the surrounding environment naturally. We introduce an object-conditioned contact generator to produce a contact feature for the human body model, which represents the contact probability for each vertex on the human body mesh. The original SMPL-X human body model has a dense (10475 vertices) and uneven vertex distribution. To address this, we adopt the mesh simplification method from POSA \cite{POSA} to downsample the vertices to 655, resulting in a more uniform distribution and reducing the number of parameters in the contact generator.

\begin{figure}[t]
  \centering
  \includegraphics[width=\linewidth]{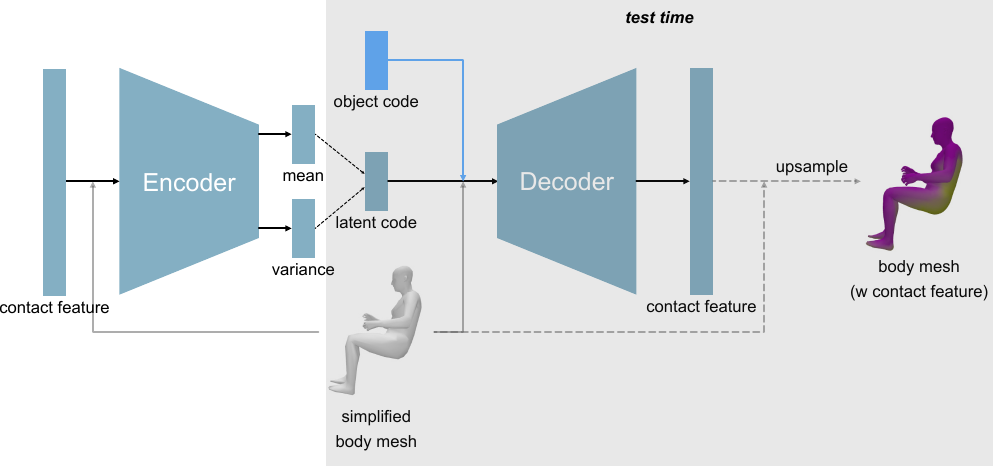}
  \caption{The network structure of the contact generator. The input and output are both the contact feature. The simplified body mesh and object code serve as the conditional input to control the generated contact feature.}
  \label{figure:contact_generator}
\end{figure}

As shown in Figure \ref{figure:contact_generator}, we utilize a VAE-based architecture to generate the body contact feature \cite{VAE, POSA}. The input contact feature $\contact \in R^{655}$ and the corresponding simplified body mesh vertices $\svertices \in R^{655 \times 3}$ are concatenated and fed into an encoder composed of spiral convolutions \cite{Spiral, SpiralPP}, which outputs the mean $\mean \in R^{256}$ and variance $\var \in R^{256}$ of the Gaussian distribution. The simplified body mesh vertices $\svertices$, sampled latent code $z \in R^{256}$, and the object code $c_o \in R^{42}$ are then fed into the decoder composed of spiral convolutions to obtain the reconstructed contact feature $\pcontact \in R^{655}$. The object code is defined as a one-hot code representing the object type and used to control the generator to produce different contact features when interacting with different objects at the same pose. During testing, we only use the decoder to generate the contact feature for a given simplified human body mesh, and then upsample the contact feature to obtain the complete contact feature.

The training loss for the contact generator can be formulated as follows:

\begin{gather}
    \cg = \recweight \recterm + \klweight \klterm \\
    \recterm = \|\pcontact - \contact\|_2^2 \\
    \klterm = \mathrm{KL}(q(Z|\svertices, \contact) \| N(0, I))
\end{gather}

\noindent $\recterm$ denotes reconstruction loss and is calculated as the mean squared error between the input and reconstructed contact feature. $\klterm$ denotes the KL divergence and is used to encourage the latent space distribution to be close to a prior distribution such as the standard normal distribution. $\lambda_{*}$ denotes the weight for each loss term.

\subsection{Putting Human into the Scene}

\subsubsection{Initial Position Selection}

After obtaining the human body model and its corresponding contact feature, the subsequent step involves placing the human body in the scene based on the contact feature. It is crucial to select appropriate initial positions and orientations to ensure interaction between the human body and the target object. Some methods utilize generative models to generate potential positions and orientations around the target object \cite{COINS}. The diversity of the generated results heavily depends on the human-scene interaction dataset. For instance, when dealing with interactions like sitting on a chair, it becomes challenging to generate humans sitting sideways at a table if the interaction dataset only includes humans sitting facing a table.

To enhance the diversity of the generated results, we employ a simple yet effective method for obtaining initial positions and orientations. Given that the scene's semantic and instance segmentation are already known, we can directly query the target objects and calculate their bounding boxes. We then uniformly sample grid points within the bounding box as initial positions above the target object. For each position, we assign four directions: front, back, left, and right. This approach enables us to obtain a diverse collection of initial positions and orientations.

\subsubsection{Physical Feasibility Test}

\begin{figure}[t]
  \centering
  \includegraphics[width=\linewidth]{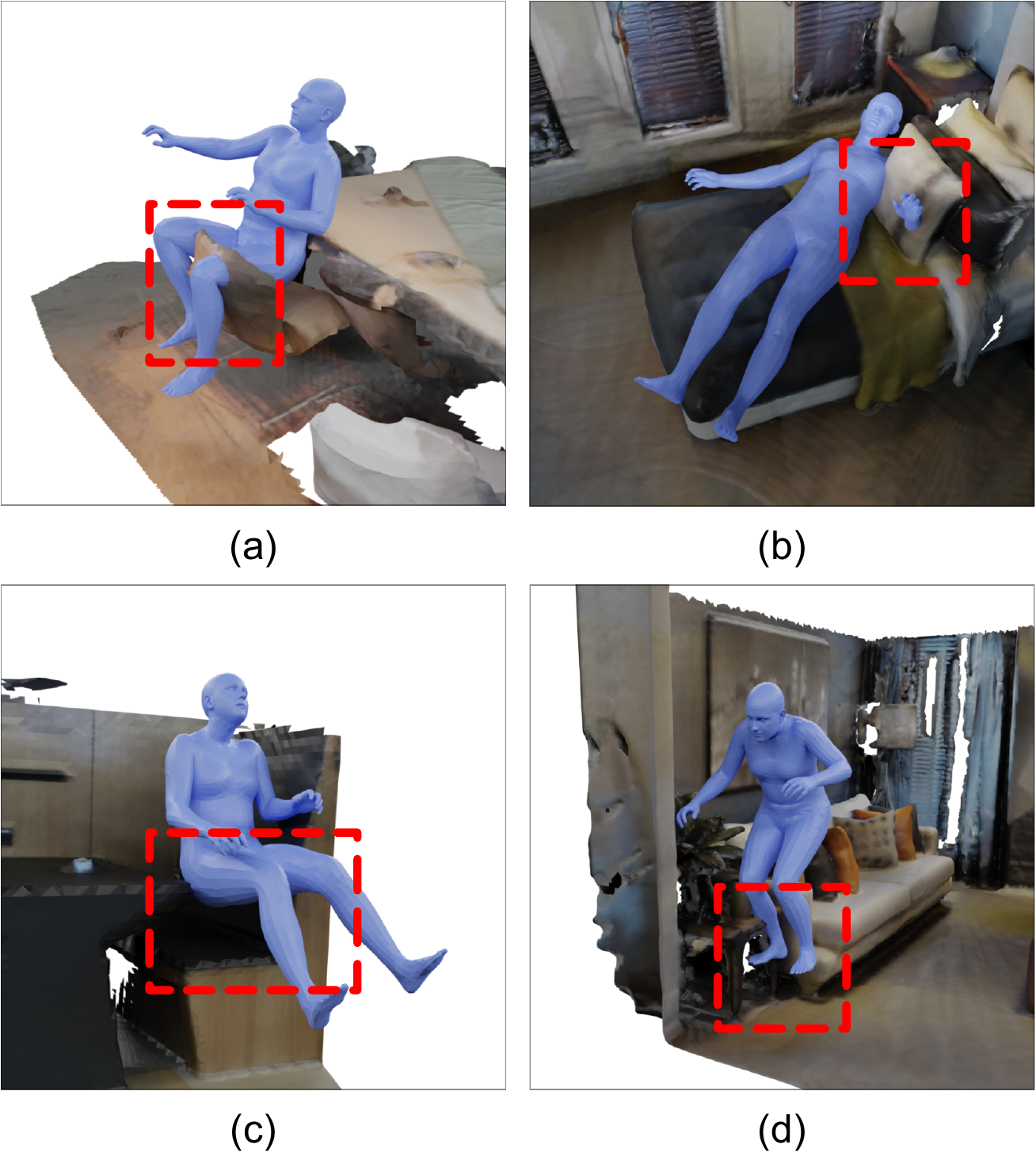}
  \caption{Examples of bad positions. (a) and (b) show the situations with severe penetrations. (c) and (d) show the situations without reasonable contact.}
  \label{figure:pft}
\end{figure}

While the above-mentioned initial positions and orientations guarantee contact between the human and the target object, they may not always yield reasonable results. Placing the human body directly at these positions may lead to issues illustrated in Figure \ref{figure:pft}. To address this concern, we introduce two physical feasibility tests including a penetration test and a contact test to eliminate unsuitable initial results.

\textbf{Penetration Test}: The penetration test is used to identify positions where the human body exhibits severe penetrations that are challenging to resolve through optimization. We refer to this type of penetration as ``thorough penetration". Figure \ref{figure:pft} (a) and (b) illustrate scenarios where the human body is divided into multiple parts due to thorough penetration. In (a), the left leg penetrates through the thin chair back, resulting in a small penetration volume that makes the penetration term too insignificant to be effectively addressed through optimization. In (b), the left hand penetrates deeply through the thick pillow, posing a significant challenge to resolve despite the large penetration term.

To detect such situations, we design a geometry-based algorithm. Specifically, we compute the scene's signed distance field (SDF) value for each vertex of the human body. We then remove faces that connect vertices with different SDF signs, which indicates the presence of penetration edges. Next, we treat the human body mesh as a graph and analyze its connected components, distinguishing between positive and negative components. If we identify two or more positive connected components, it suggests the presence of thorough penetrations, and we eliminate the corresponding position from consideration.

\textbf{Contact Test}: The contact test is employed to identify situations where the human body lacks ``reasonable contact" with the scene. It is important for the human body to have support from the scene in order to ensure physically plausible interactions. In Figure \ref{figure:pft} (c) and (d), even though certain body parts, such as the left arm and left leg in (c), or the right leg in (d), are close to the scene, the body parts that should be in contact with the scene are not actually making contact. For example, in a standing pose, the soles of the feet should be in contact; in a sitting pose, the thighs; and in a lying pose, the back, thighs, and legs.

To detect these situations, we propose using the contact feature generated in the last stage. We define the real contact vertices as those vertices that not only have a high contact probability but are also close to the target object. For each position, we calculate the number of real contact body vertices and remove positions where this number falls below a certain threshold. This ensures that only positions with a sufficient number of body vertices in genuine contact with the target object are retained.

\subsubsection{Optimization}

Although the penetration test and the contact test eliminate most unreasonable results, some slight penetrations or lack of necessary contact may still exist in the initial results. To further enhance the realism of the results, we additionally optimize the human pose. The objective function is defined as follows:

\begin{equation}
    E = \wcweight \wcterm + \vpweight \vpterm + \rweight \rterm
\end{equation}

\noindent $\wcterm$ denotes the weighted contact term and is used to enforce necessary contact between the human body and the target object. $\vpterm$ denotes the volume penetration term and is used to reduce the penetrations between the human body and the scene. $\rterm$ is the regularization term to minimize the mean squared error between the current body pose and the initial body pose. $\lambda_{*}$ denotes the weight for each term.

\textbf{Weighted Contact Term}: We consider body vertices with contact probability above a certain threshold as contact vertices $\cvertices$. We sample points on the target object to get object points $\opoints$. Then, we minimize the weighted distances from $\cvertices$ to $\opoints$ to match the human body with the target object:

\begin{equation}
    \wcterm = \sum\limits_{v_i \in \cvertices} f_i \rho(\min_{p_j \in \opoints} \|v_i - p_j\|_2)
\end{equation}

\noindent $\rho(\cdot)$ denotes a robust Geman-McClure error function \cite{GemanMcClure} for down weighting the vertices in $\cvertices$ that are far from $\opoints$. $f_i$ denotes the contact probability of contact vertex $v_i$. 

\textbf{Volume Penetration Term}: We extract the human body internal points $\internal$ by treating the human body as a volume \cite{OcclusionHPR}. Then we minimize the sum of the absolute values of the scene SDF for internal points with a negative scene SDF $\internal^{-}$ to reduce penetrations between the human body and the scene:

\begin{equation}
    \vpterm = \sum\limits_{p_i \in \internal^{-}} |\mathrm{SDF}(p_i)|
\end{equation}

\noindent $\mathrm{SDF}(\cdot)$ denotes the function to search for the scene SDF for a given point.

%% file: sections/4-experiment.tex
\section{Experiments}

\subsection{Datasets}

The dataset we used consists of three parts: the human dataset (e.g. AMASS \cite{AMASS}), the human-scene interaction dataset (e.g. PROX \cite{PROX}), and the scene dataset (e.g. MP3D-R \cite{MP3D, PSI}). 

\textbf{AMASS}: We utilize the AMASS dataset to train the pose generator. BABEL dataset \cite{BABEL} provides sequence-based and frame-based annotations, enabling us to train the pose generator with action conditioning. Specifically, we use 4 subsets including ACCAD, HDM05, CMU, and BMLrub.

\textbf{PROX}: We leverage the PROX dataset \cite{PROX} to train both the pose generator and the contact generator. Following the split of PSI \cite{PSI}, we use 8 scenes as the training set and 4 scenes as the testing set.

\textbf{MP3D-R}: To further evaluate the generalization ability of our method, we also test our approach on scenes from the MP3D-R dataset \cite{MP3D, PSI}.

\begin{figure*}[htbp]
  \centering
  \includegraphics[width=\linewidth]{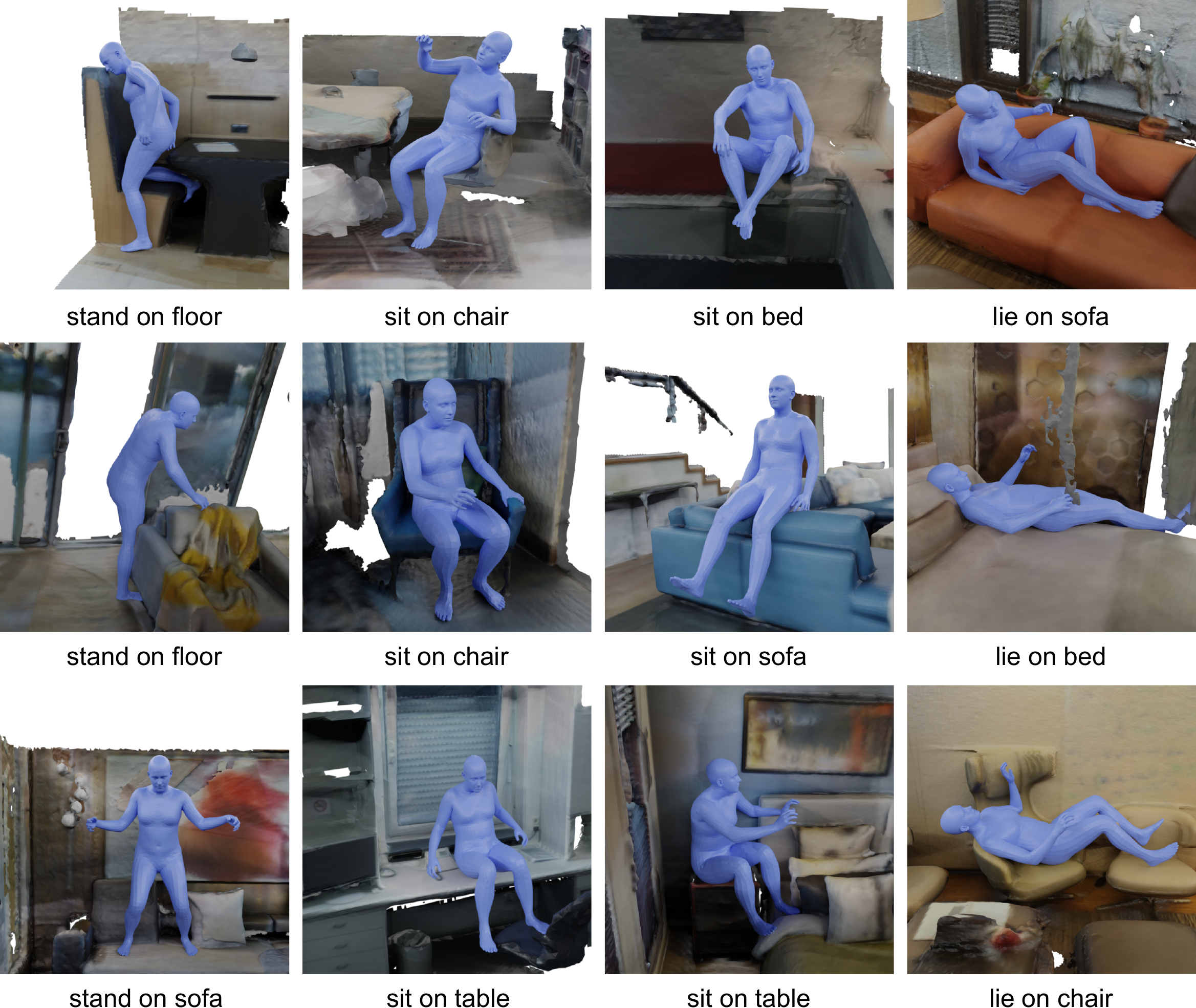}
  \caption{Gallery of our results. The first and second row denotes results on the PROX and MP3D-R dataset respectively. The last row denotes results under uncommon interactions.}
  \label{figure:ours}
\end{figure*}

\subsection{Experiment Details}

\textbf{Pose Generator}: We train the pose generator using the Adam optimizer \cite{Adam} with a learning rate of 1e-3. The weights for each loss term are set as follows: $\mweight=2$, $\vweight=4$, $\jweight=2$, and $\klweight=0.005$.

\textbf{Contact Generator}: We create the training data for the contact generator using LEMO \cite{LEMO} fitting. For each body vertex, we calculate the ground truth contact probability as $\mathrm{Clamp}(1-\frac{d}{\delta})$, where $d$ denotes the minimum distance from the body vertex to the scene, and $\delta$ is the distance threshold set to 0.05. The $\mathrm{Clamp}(\cdot)$ function clamps the value to $0\sim1$. We train the contact generator using the Adam optimizer with a learning rate of 1e-3. The weights for each loss term are set as follows: $\recweight=1$, and $\klweight=0.001$.

\textbf{Optimization}: We employ the same optimizer as SMPLify-X \cite{SMPLX}. The weights for each term are set as follows: $\wcweight=1$, $\vpweight=10$, and $\rweight=50$.

\begin{figure*}[htbp]
  \centering
  \includegraphics[width=\linewidth]{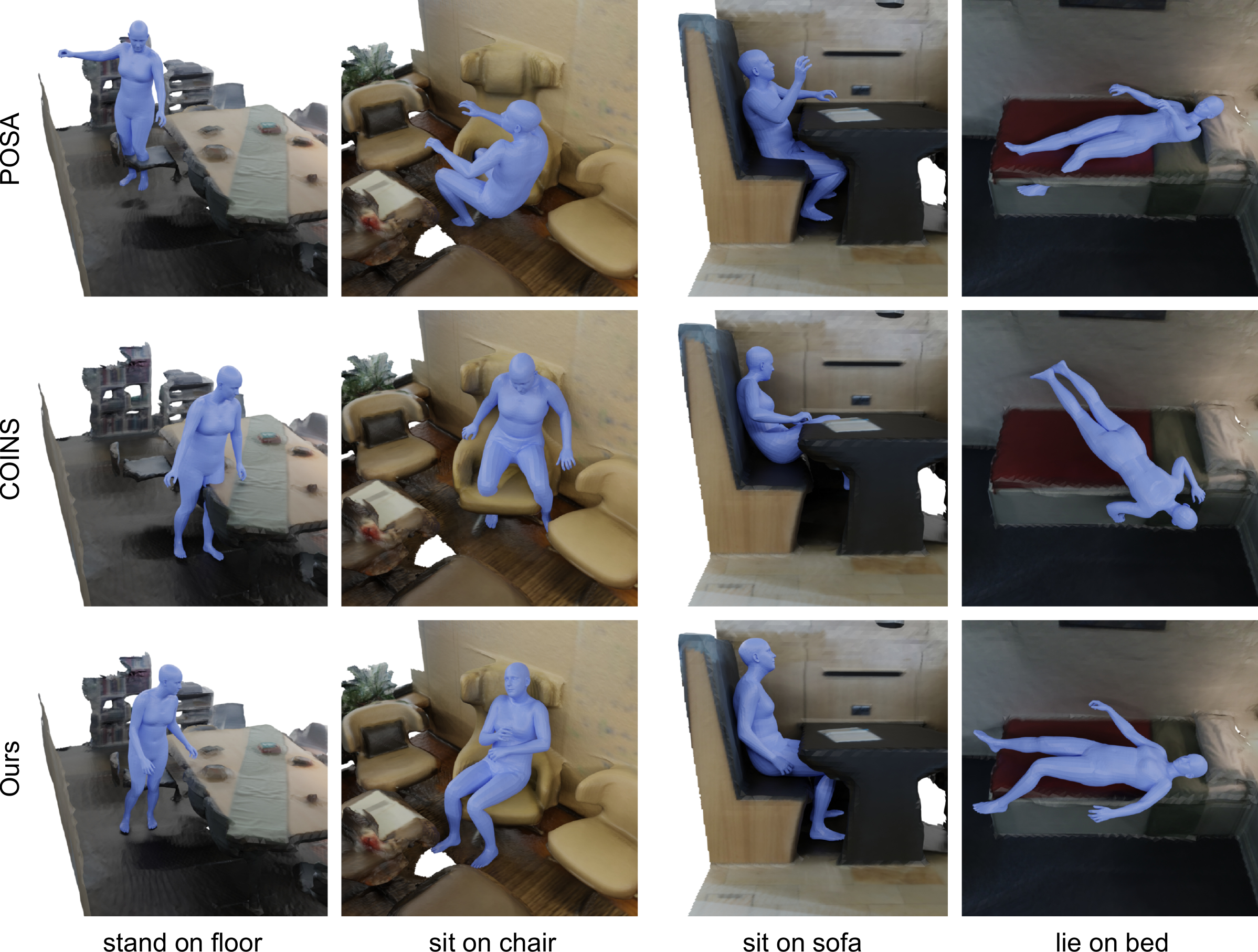}
  \caption{Comparison of the generated results by our method with those of POSA and COINS.}
  \label{figure:comp}
\end{figure*}

\subsection{Results}

Figure \ref{figure:ours} presents a gallery of our results. We test our method on some common interactions that appear frequently in the dataset, such as standing on the floor, sitting on the chair, sitting on the sofa, or lying on the bed. The first and second rows show results on the PROX and MP3D-R dataset respectively. We can see that our method can generate physically plausible interactions for different action-object pairs. By decoupling the pose and interaction process, our method can generate some uncommon body poses, such as bending the knees (row 1, column 1) or crossing the legs (row 1, column 4). The contact generator can ensure necessary contact so the human body looks more natural without feeling isolated from the scene (row 2, column 1-3). It should be noted that the strip-shaped mesh above the human body is part of the original scene mesh (row 2, column 4).

In the third row of Figure \ref{figure:ours}, we test our method on some uncommon interactions that never appear or seldom appear in the dataset, such as standing on the sofa, sitting on the table, or lying on the chair, to further validate the generalization ability of our method. We can see that our method can still generate reasonable interactions.

\subsection{Comparison and Evaluation}

In this section, we present the comparison results of our method with other methods on the PROX dataset.

\subsubsection{Evaluation Metrics}

We evaluate the performance using a set of metrics, which can be categorized into physical plausibility and pose diversity metrics. 

\textbf{Physical Plausibility}: To evaluate the physical plausibility, we employ the Non-Collision (NC) and Contact metric introduced in PSI \cite{PSI}. NC is used to measure the penetration and is calculated as the ratio of body vertices with a positive scene SDF value. Contact is 1 if any body vertex has a negative scene SDF value, otherwise, it will be 0. We additionally use the Volume Non-Collision (VNC) to compensate for the defect of NC \cite{OcclusionHPR}. 

\textbf{Pose Diversity}: Following PSI, we cluster the body poses into 50 clusters using K-Means \cite{KMeans}. Then we calculate two metrics to evaluate the diversity. The first is the entropy of the cluster-ID histogram. It measures the average degree of all clusters. The second is the cluster size which is calculated as the average distance between the cluster center and the samples belonging to it. It measures the diversity degree of each cluster.

\subsubsection{Comparison Results}

We compare our method with POSA \cite{POSA} and COINS \cite{COINS}. To enable POSA to support the same input, we randomly select corresponding poses from the PROX dataset and sample initial positions around the object. The comparison results are listed in Table \ref{table:results}. For the physical plausibility metrics, the penetration and contact metrics are conflicting, making it challenging to strike a balance. Nevertheless, our method achieves the highest value for all metrics. In terms of pose diversity metrics, the Entropy metric shows no significant differences among all methods. However, our method demonstrates a notable improvement on the Cluster Size metric, indicating that our pose generator learns richer pose prior.

\begin{table}[htbp]
    \centering
    \small
    \caption{Comparison results on the PROX dataset.}
    \resizebox{\linewidth}{!}{
    \begin{tabular}{l|ccc|cc}
    \toprule
        & \multicolumn{3}{c|}{Physical Plausibility} & \multicolumn{2}{c}{Diversity} \\
    \cmidrule{2-6}
        & NC $\uparrow$ & VNC $\uparrow$ & Contact $\uparrow$ & Entropy $\uparrow$ & Cluster Size $\uparrow$ \\
    \midrule
        POSA & 0.96 & 0.94 & 0.95 & 3.60 & 0.68 \\
        COINS & \textbf{0.99} & 0.98 & 0.89 & \textbf{3.77} & 0.61 \\
        \textbf{Ours} & \textbf{0.99} & \textbf{0.99} & \textbf{0.96} & 3.69 & \textbf{0.90} \\
    \bottomrule
    \end{tabular}}
    \label{table:results}
\end{table}

In Figure \ref{figure:comp}, we visually compare our method with other methods on some common interactions. In column 1, 2, and 4, other methods face the problem of severe penetrations or no contact, while our method can avoid such situations due to the physical feasibility test module. In scenes with constrained space, it is hard to put the human body at a suitable place with no penetrations at one time, such as the sofa in column 3. Our method can adjust the human pose to reduce the penetrations using the optimization module.

\begin{table*}[t]
    \centering
    \small
    \caption{Evaluation based on different actions.}
    \resizebox{0.8\linewidth}{!}{
    \begin{tabular}{l|ccc|ccc|ccc}
    \toprule
        & \multicolumn{3}{c|}{Stand} & \multicolumn{3}{c|}{Sit} & \multicolumn{3}{c}{Lie} \\
    \cmidrule{2-10}
        & NC $\uparrow$ & VNC $\uparrow$ & Contact $\uparrow$ & NC $\uparrow$ & VNC $\uparrow$ & Contact $\uparrow$ & NC $\uparrow$ & VNC $\uparrow$ & Contact $\uparrow$ \\
    \midrule
        POSA & 0.97 & 0.98 & 0.87 & 0.96 & 0.93 & 0.99 & 0.96 & 0.93 & 0.97 \\
        COINS & 0.98 & 0.99 & 0.80 & \textbf{0.99} & 0.97 & \textbf{1.00} & \textbf{0.99} & \textbf{0.99} & 0.65 \\
        \textbf{Ours} & \textbf{0.99} & \textbf{1.00} & \textbf{0.90} & \textbf{0.99} & \textbf{0.99} & 0.98 & \textbf{0.99} & \textbf{0.99} & \textbf{1.00} \\
    \bottomrule
    \end{tabular}}
    \label{table:results_action}
\end{table*}

We further analyze the results by actions in Table \ref{table:results_action}. Our method outperforms others in all metrics, except for the Contact metric of the sitting pose. For the standing pose, since the human body only contacts the scene with the soles of the feet, it is easy to encounter floating issues. In contrast, sitting or lying poses involve more contact with the scene, which can lead to penetrations. Achieving a balance between less penetration and more contact is a challenging task. However, our method successfully strikes a balance for different actions.

\subsection{Ablation Study}

We conduct the ablation study on the PROX dataset. We consider the following ablation versions:

\begin{itemize}
  \item Ours (w/o PFT): we test how our method performs when the Physical Feasibility Test module is removed.
  \item Ours (w/o OPT): we test how our method performs when the OPTimization module is removed.
\end{itemize}

\noindent As shown in Table \ref{table:ablation}, the final version which includes all the proposed components achieves the best overall performance.

\begin{table}[htbp]
    \centering
    \small
    \caption{Ablation study on the PROX dataset.}
    \resizebox{\linewidth}{!}{
    \begin{tabular}{l|cc|ccc}
    \toprule
        & PFT & OPT & NC $\uparrow$ & VNC $\uparrow$ & Contact $\uparrow$ \\
    \midrule
        Ours (w/o PFT) &  & \checkmark  & 0.97 & 0.96 & 0.95 \\
        Ours (w/o OPT) & \checkmark &  & \textbf{0.99} & 0.98 & 0.87 \\
        \textbf{Ours} & \checkmark & \checkmark & \textbf{0.99} & \textbf{0.99} & \textbf{0.96} \\
    \bottomrule
    \end{tabular}}
    \label{table:ablation}
\end{table}

\begin{figure}[htbp]
  \centering
  \includegraphics[width=\linewidth]{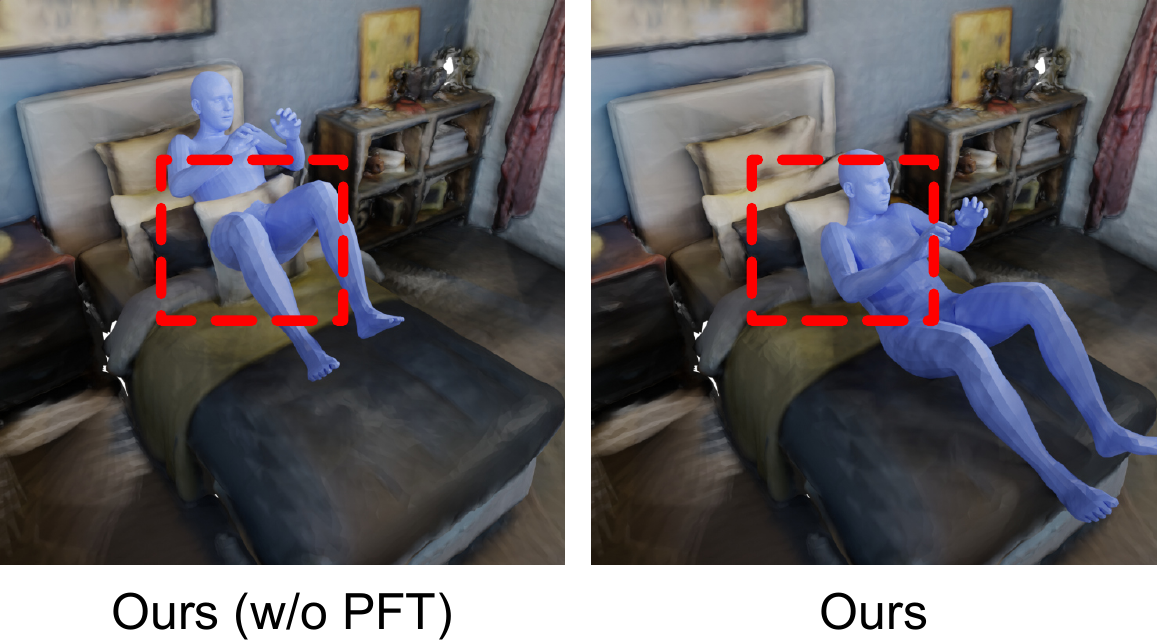}
  \caption{Ablation study: compare our method with the version without physical feasibility test.}
  \label{figure:ablation_pft}
\end{figure}

Without the physical feasibility test module, severe penetrations may occur in the initial results, leading to a decline in the NC and VNC metrics. As shown in Figure \ref{figure:ablation_pft}, the physical feasibility test module enables us to avoid undesirable results, such as the human body penetrating through a pillow, thereby ensuring more realistic and physically plausible interactions.

\begin{figure}[htbp]
  \centering
  \includegraphics[width=\linewidth]{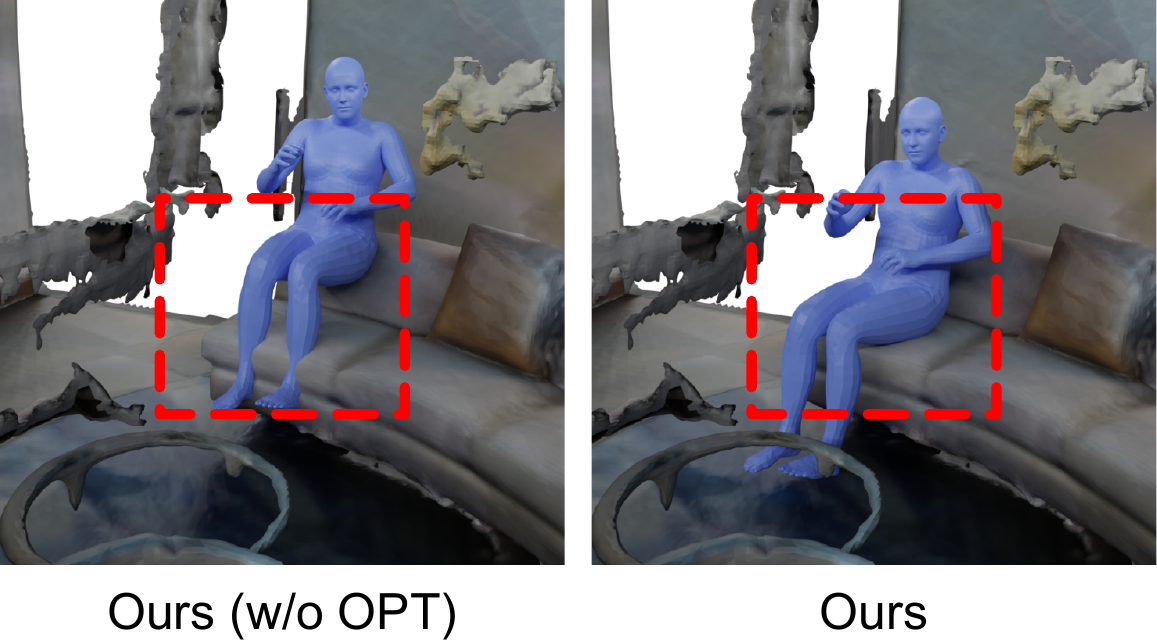}
  \caption{Ablation study: compare our method with the version without optimization.}
  \label{figure:ablation_opt}
\end{figure}

When the optimization module is removed, the direct placing result may exhibit penetrations or fail to establish required contact with the scene, resulting in a significant decline in the Contact metric. As illustrated in Figure \ref{figure:ablation_opt}, even with a suboptimal initial result, the optimization module can effectively guide the human to sit in a comfortable and natural manner.

%% file: sections/5-conclusion.tex
\section{Conclusion}

We present a novel method for generating diverse 3D human poses in scenes with semantic control. Our main idea is to decouple the pose and interaction generation process so that we can minimize the reliance on the human-scene interaction dataset. This decoupled structure enables us to learn a richer pose prior, resulting in more diverse human poses. We introduce a physical feasibility test module to eliminate undesirable positions, avoiding severe penetrations or lack of reasonable contact. An optimization module is also proposed to fine-tune the human pose, making the result appear more natural. Extensive experiments demonstrate that our method can generate more physically plausible interactions with more diverse human poses compared to other methods.

\textbf{Limitations and future work}: Our method can only handle fixed text descriptions in the form of ``action + object", which restricts its usability. A potential direction for future work is to integrate language models that can extract key control information from complex descriptions. This will provide more precise control over the generated results. Furthermore, our experiments are confined to indoor scene datasets, whereas outdoor scenes present distinct object categories and structures with different human-scene interactions. Future work can extend 3D human generation to outdoor scenes to enhance the generality.

%% file: sections/6-acknowledgments.tex
\section*{Acknowledgements}

This work was supported by the National Natural Science Foundation of China (62072366, U23A20312) and National Key R\&D Program of China (2022YFB3303200).